%% file: root.tex
\documentclass[letterpaper, 10 pt, conference]{ieeeconf}  % Comment this line out if you need a4paper

\IEEEoverridecommandlockouts                              % This command is only needed if 
\usepackage{amsmath}
\usepackage{amssymb}
\usepackage{graphicx}
\usepackage{float}
\usepackage{array}
\usepackage{tikz}
\usepackage{listings}
\usepackage{mathrsfs}
\usepackage{pgfplots}
\usepackage{python}
\usepackage[pdfencoding=auto,psdextra,pagebackref,breaklinks,colorlinks]{hyperref}
\usepackage{graphicx}
\usepackage{cite}
\usepackage{dsfont}
\usepackage{mathtools,etoolbox}
\usepackage[none]{hyphenat}
\usepackage[utf8]{inputenc}
\usepackage[english]{babel}

\usepackage{amsthm}
\usepackage{xfrac}
\usepackage{nicefrac}
\usepackage{booktabs}
\usepackage[switch, pagewise]{lineno}
\usepackage{nicefrac}
\usepackage{flushend}
\usepackage{gensymb}
\newcommand\numberthis{\addtocounter{equation}{1}\tag{\theequation}}
\title{\LARGE \bf
Detecting Olives with Synthetic or Real Data? Olive the Above}
\author{Yianni Karabatis$^{1}$, Xiaomin Lin$^{1}$, Nitin J. Sanket$^{2}$, Michail G. Lagoudakis$^{3}$, Yiannis Aloimonos$^{1}$ %
\thanks{*This work was funded by the Fulbright foundation}% <-this % stops a space
\thanks{$^{1}$Perception and Robotics Group, University of Maryland Institute for Advanced Computer Studies, University of Maryland, College Park, MD 20742, USA. Emails: \texttt{\{yianni,xlin01,jyaloimo\}@umd.edu}.
        }%
\thanks{$^{2}$Perception and Autonomous Robotics Group, Robotics Engineering, Worcester Polytechnic Institute, MA 01609, USA. Email: \texttt{nsanket@wpi.edu}.%
}
\thanks{$^{3}$School of ECE, Technical University of Crete, Greece. Email: \texttt{lagoudakis@tuc.gr}.%
}
}

\begin{document}

\maketitle
\thispagestyle{empty}
\pagestyle{empty}

%%%%%%%%%%%%%%%%%%%%%%%%%%%%%%%%%%%%%%%%%%%%%%%%%%%%%%%%%%%%%%%%%%%%%%%%%%%%%%%%
\input{sections/0.abstract}
\input{sections/1.introduction}
\input{sections/2.related_work}
\input{sections/3.proposed_method}
\input{sections/4.Experiments_And_Results}
\input{sections/5.conclusions}

%%%%%%%%%%%%%%%%%%%%%%%%%%%%%%%%%%%%%%%%%%%%%%%%%%%%%%%%%%%%%%%%%%%%%%%%%%%%%%%%

% \addtolength{\textheight}{-12cm}   % This command serves to balance the column lengths
                                  % on the last page of the document manually. It shortens
                                  % the textheight of the last page by a suitable amount.
                                  % This command does not take effect until the next page
                                  % so it should come on the page before the last. Make
                                  % sure that you do not shorten the textheight too much.

%%%%%%%%%%%%%%%%%%%%%%%%%%%%%%%%%%%%%%%%%%%%%%%%%%%%%%%%%%%%%%%%%%%%%%%%%%%%%%%%
% \section*{APPENDIX}

% Appendixes should appear before the acknowledgment.

\section*{ACKNOWLEDGMENTS}

This work was partially sponsored by the Fulbright Foundation and USDA NIFA Award\# 20206801231805. We thank all olive grove farmers, Dr. Panagiotis Partsinevelos' SenseLab, and Zisis Charokopos for data collection assistance.

%%%%%%%%%%%%%%%%%%%%%%%%%%%%%%%%%%%%%%%%%%%%%%%%%%%%%%%%%%%%%%%%%%%%%%%%%%%%%%%%

\bibliographystyle{IEEEtran}
\bibliography{IEEEabrv,refs}

\end{document}

%% file: sections/0.abstract.tex
\begin{abstract}

%Something about climate change, diseases, the emergence of precision agriculture, how vital yield estimation is, synthetic data, world's only olive dataset.
Modern robotics has enabled the advancement in yield estimation for precision agriculture. However, when applied to the olive industry, the high variation of olive colors and their similarity to the background leaf canopy presents a challenge. Labeling several thousands of very dense olive grove images for segmentation is a labor-intensive task. This paper presents a novel approach to detecting olives without the need to manually label data. In this work, we present the world’s first olive detection dataset comprised of synthetic and real olive tree images. This is accomplished by generating an auto-labeled photorealistic 3D model of an olive tree. Its geometry is then simplified for lightweight rendering purposes. In addition, experiments are conducted with a mix of synthetically generated and real images, yielding an improvement of up to 66\% compared to when only using a small sample of real data. When access to real, human-labeled data is limited, a combination of mostly synthetic data and a small amount of real data can enhance olive detection.
\end{abstract}

%% file: sections/1.introduction.tex
\section{INTRODUCTION}
\label{section:introduction}

The olive tree, native to the Mediterranean region, has played an essential role in the advancement of humanity for millennia. The olive harvest is a lengthy process requiring almost a year-long preparation and is essential for the global economy, with a market valued at \$13.77B and is forecasted to reach \$17.99B by 2029 \cite{fortune_business_insights_2022}.

%However, climate change has negatively affected olive-growing regions, thereby resulting in a stronger disease proliferation.Hence, it is imperative that precision agriculture methods account for pre-emptive disease prognoses and additionally for the yield estimation of at-risk crops.

Yield estimation aims to forecast the output of crops. Manual inspections help farmers better understand their crops, but patrolling a grove with thousands of trees is time-consuming. One solution is to utilize Unmanned Aerial Vehicles (UAVs) to capture and analyze aerial image data for making better decisions.
% \begin{figure*}[ht!]
% \includegraphics[width=0.9\textwidth]{./figures/visual_comparison.pdf}
% %\includegraphics[width=\textwidth,height=\textheight,keepaspectratio]{sequence.png}
% \centering
% \caption{The results of our approach. column 1 is the input data. columns 2-4 show the respective segmentations without synthetic data, with synthetic data, and with synthetic data in the IGA space]}
% \label{fig:visual_comparison}
% \end{figure*}
\begin{figure}
 \centering
%{\includegraphics[width=1.0\linewidth]{./figures/visual_comparison_v3.pdf}}
{\includegraphics[width=1.0\linewidth]{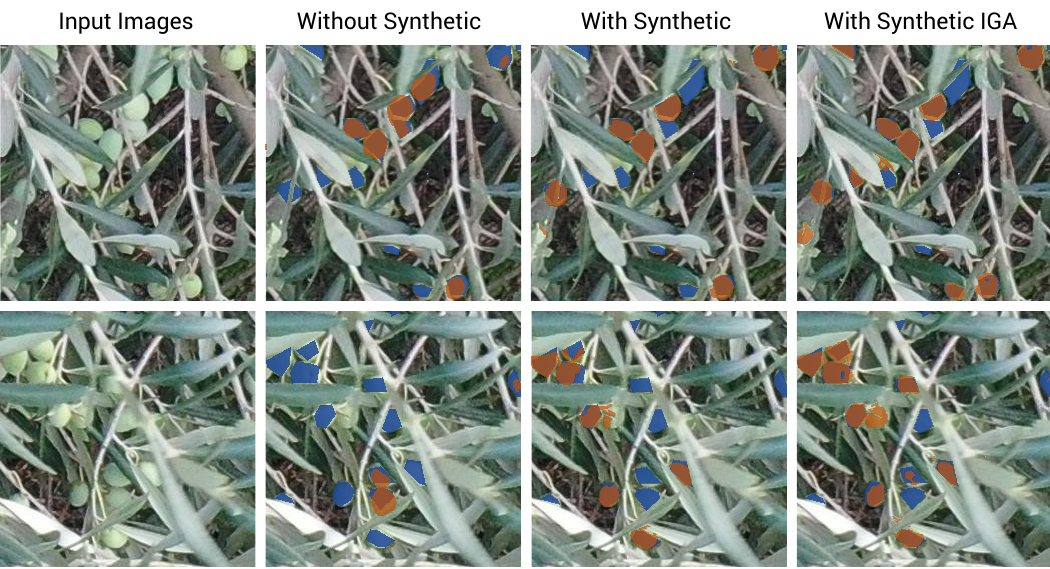}}
\vspace{-5mm}
\caption{Each column left to right: Input image, prediction using only real data, prediction using real and synthetic data, prediction using real and synthetic data in IGA. Predictions are shown in {\color[rgb]{0.8,0.3,0.1}orange} and ground truths are shown in {\color[rgb]{0.1,0.3,0.9}blue}. Adding synthetic data to the training set increases the number of correct predictions.}
\vspace{-5mm}
\label{fig:detection_result}
\end{figure}

Object detection via robotics has been employed for multiple types of crops \cite{chen2017counting, crop_gen,soybean,state_aware_tracker,liu2022information}. However, olives present additional problems not encountered with other crops. Specifically, one must account for moderate variation in olive color, since the olive may have multiple shades of green. Another problem is the similarity between the olive fruit and the surrounding leaves, making it challenging to differentiate between the two. This similarity can be exacerbated even further when only the end of an olive is visible, resembling a leaf. In addition, the dense canopy of an olive tree occludes many of the olives, making them partially visible and difficult to recognize as shown in Fig.~\ref{fig:detection_result}. Finally, the average size of an olive is $2\text{cm} \times 1\text{cm} \times 1\text{cm} $, magnifying all the difficulties mentioned above.

%We then propose a high-performing, automated method of detecting olives in olive groves to aid farmers in making better-informed crop-management decisions.
\begin{figure*}[ht!]
\includegraphics[width=0.9\textwidth]{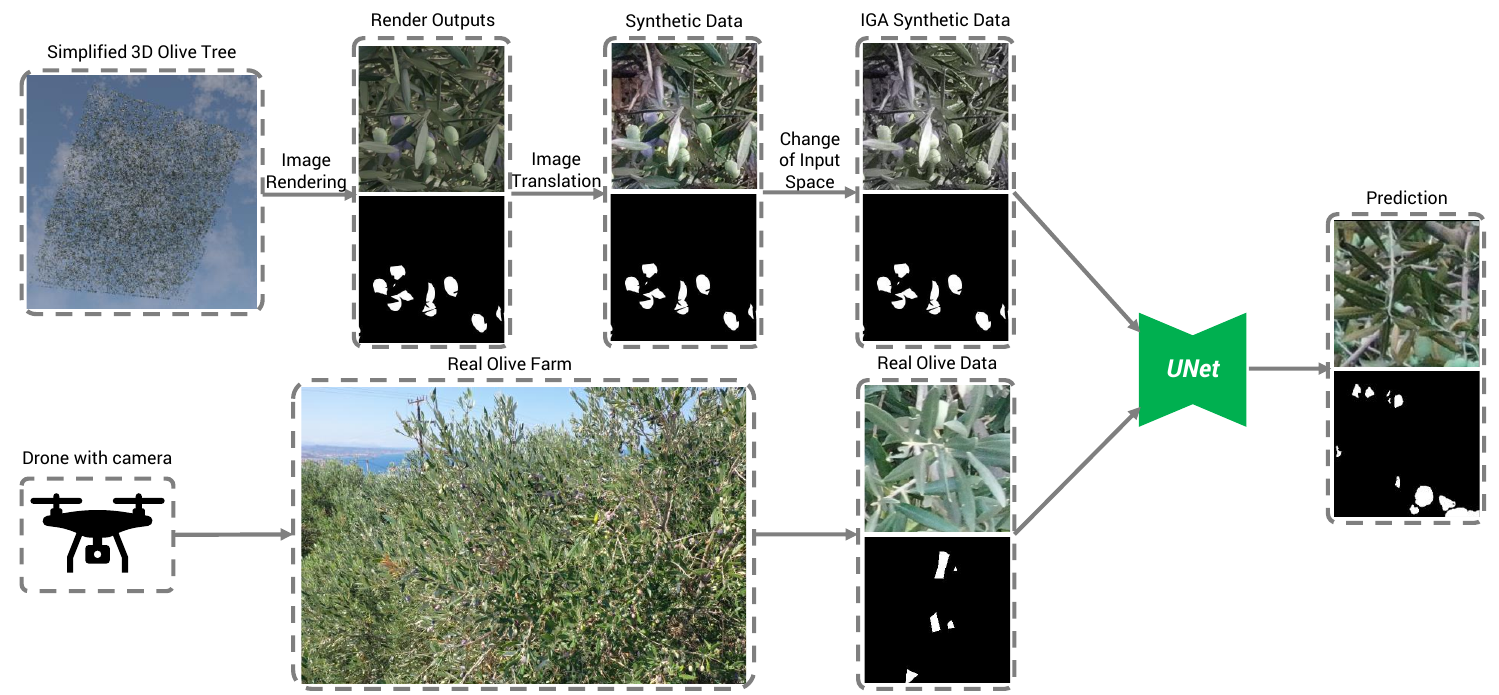}
\centering
\caption{An overview of our approach: The proposed geometric model is used to generate synthetic images which are further fed into a VSAIT to enable image-to-image translation. We then combine the synthetic data with real data to train an Unet for olive detection.}
\label{fig:overview}
\vspace{-3mm}
\end{figure*}
However, labeling images collected from UAVs is an exhausting task, especially when there could be up to 17,500 olives\cite{dz_2022} in a single olive tree (up to 1000 olives in a single image from our observation). An olive grove may contain thousands of olive trees, making this very labor-intensive. {\em Therefore, we investigate a new approach to cover the deficiency of labeled data, namely we create and incorporate synthetic data as an aid to training models for olive segmentation}. Synthetic data is automatically labeled and can be easily scaled up to higher orders of magnitude. We create synthetically generated image-mask pairs of an olive tree, including its olives, leaves, branches, and surrounding background. We propose the use of our synthetic data to create a high-performing, automated system for detecting olives in olive groves. The contributions of this paper are as follows:
\begin{itemize}
%\item We propose a novel mathematical model for the 3D shape of olives.
\item We create the world's first olive detection dataset consisting of mostly synthetic and some real images of olive trees
\begin{itemize}
    \item We generate a photorealistic 3D model of an olive tree to create an auto-labeled synthetic dataset
    \item We propose and simplify a geometric model for olives on an olive tree for lightweight rendering purposes
\end{itemize}
\item We create a novel color input space for our synthetic data, making it more generalizable to real-world data
\item We experiment with segmentation models to assist in situations where the amount of real labeled data is very limited
\end{itemize}
The remainder of this paper is structured as follows: Section ~\ref{section:related_work} presents related works. In Section~\ref{section:Synthetic_Olive_Tree} we describe the generation of the 3D synthetic olive tree model and discuss the image-to-image translation technique used to create realistic synthetic images.
 Section.~\ref{section:Experiments_and_results} presents the experiments using a mix of synthetic and real data. Finally, Section.~\ref{section:Conclusion} contains the conclusion and future work.

% 1. The world's first labeled olive detection dataset
% 1.a) Labeled real data
% 1.b) Simplify the geometric modeling of an olive tree for lightweight rendering to generate auto-labeled photorealistic synthetic images of olives in an olive tree.
% 2.) Olive detection system where synthetic data outperforms real data.
%Modern methods for yield estimation utilize robotics to their advantage as it may automate the process in a faster, less tiring way. 

%% file: sections/2.related_work.tex
\section{RELATED WORK}
\label{section:related_work}
%\colorbox{yellow}{Problem: population of Earth increases, demand of food, maximize fertility of land (arable), climate change, diseases.}
%The Earth faces many unprecedented issues today due to the human rise in population. Such an increase will correspondingly propel the demand for global food production while simultaneously amplifying food shortages.

Climate change along with the rise of the human population calls for efficient and modern agricultural solutions \cite{elijah_2023}. Precision agriculture offers a partial solution by maximizing crop outputs while minimizing the environmental footprint \cite{balafoutis2017precision}. %Recent advancements in robotics have played a pivotal role in precision agriculture, search and rescue missions and livestock monitoring \cite{alanezi2022livestock,search_rescue}.  %Moreover, it has enabled farmers to reduce costs and treat crops more efficiently using modern robotics \cite{elijah_2023,balafoutis2017precision}.

UAVs have reduced the time required to monitor agricultural properties \cite{livanos2020extraction}. However, RGB cameras cannot detect phenomena invisible to the eye, such as early symptoms of plant diseases. Therefore, UAVs have been armed with thermal, multispectral, and hyperspectral cameras that see beyond the human eye \cite{hyperspectralsurvey} to calculate certain vegetation indices (NDVI, NDRE, CWSI, etc.) and diagnose diseases early \cite{POBLETE2021133,halk}.

UAVs are also utilized to track and monitor livestock and perform search and rescue missions \cite{search_rescue}. Alanezi \textit{et al.} discuss tracking animals \cite{alanezi2022livestock}. However, the number of objects involved in livestock detection is fewer than in tracking fruit production.
 \begin{figure}[t]
\includegraphics[width=0.9\linewidth]{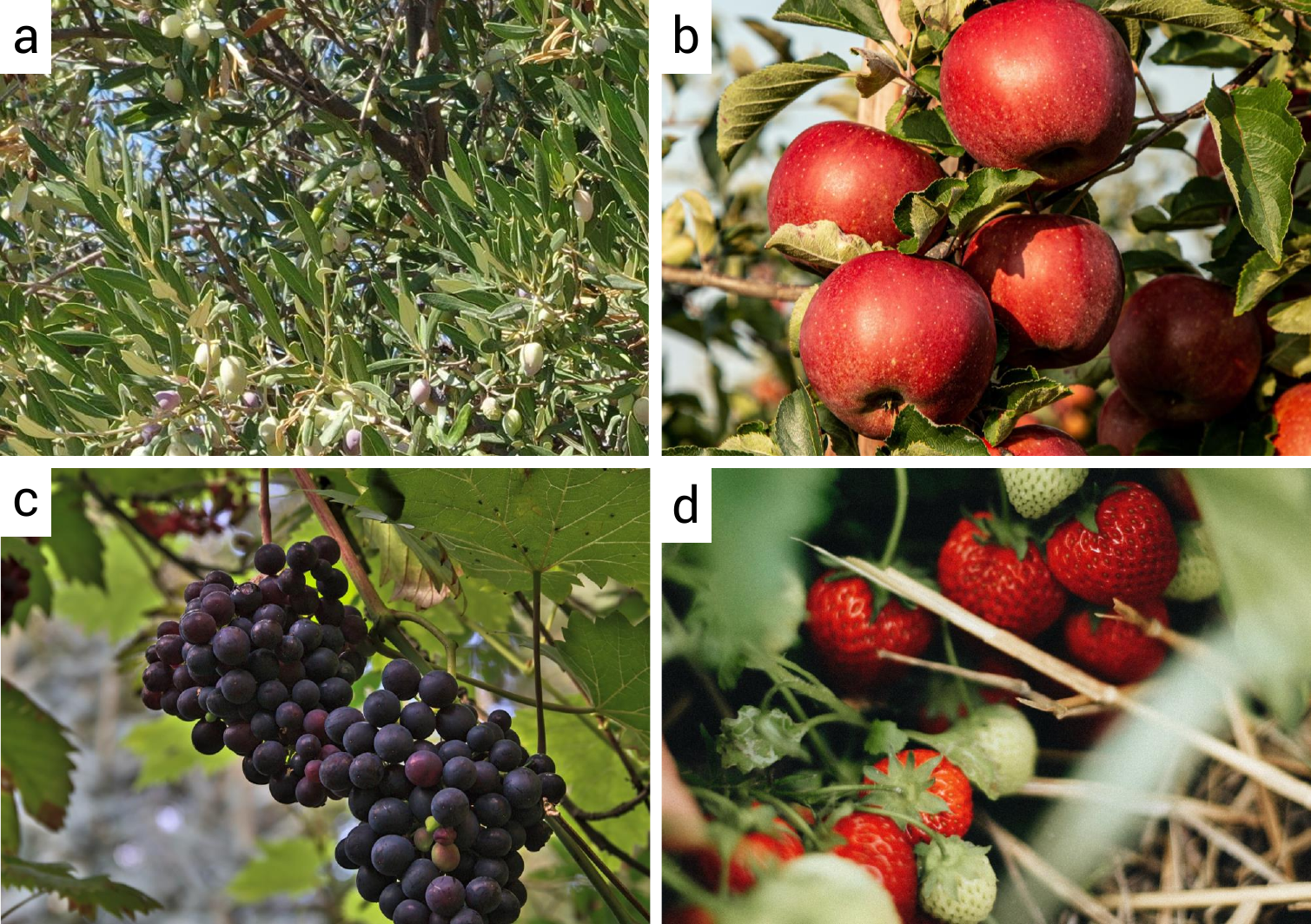}
\centering
\caption{(a): Olive Tree. (b): Apple Tree. (c): grape Tree. (d): Strawberry Bush. Olives are small and they have the same color as the background. Some fruits stand out among their background given their large size, and color differences.}
\label{fig:comparison}
\end{figure}

 Robotics coupled with machine learning techniques has assisted in better segmentation of crops \cite{weed,su2021data}. Chen \textit{et al.} used deep learning techniques to count apples and oranges in orchards given UAV-captured data \cite{chen2017counting}. Similar approaches have been utilized to identify apples, bananas, grapes (individually and in clusters), pears, and pineapples \cite{minneapple,pear,Pineapple,bananas,grapeslam,grapecluster}. However, these fruits are easy to identify given their large size, and their striking color differences against their respective backgrounds (Figs.~\hyperref[fig:comparison]{3b, 3c, 3d}). Detecting fruits of small size is even more of a challenge when the leaves partially obstruct the fruits. Computer vision efforts to count fruits of small sizes, such as cranberries, blueberries, and cherries have been explored in \cite{cranberries,blueberries,cherries}. Olives are of small sizes, making them more challenging to identify as we can see in Fig.~\hyperref[fig:comparison]{3a}. Furthermore, they also blend very well into the surrounding background which has a very similar color, making detection very challenging.

The closest research to our proposed olive detection method is systems to count how many olive trees there are in a grove \cite{gonzalez2007applying,khan2018remote,ponce2019automatic}. Ponce \textit{et al.} proposed a direct method to count olives however only after they are harvested and sent for processing in an external image acquisition chamber \cite{ponce2019automatic}. Therefore, to the best of our knowledge, we propose the first autonomous on-site olive detection system.

Previous approaches utilize synthetic data to detect objects for both agricultural and non-agricultural purposes. Lin \textit{et al.} \cite{lin2023seadronesim} uses the 3D model of a BlueROV and creates photo-realistic aerial image datasets with the ground truth for BlueROV detection. Sanket \textit{et al.} \cite{sanket2021evpropnet} model the 3D geometry of a quadrotor propeller to generate a vast, automatically-labeled dataset which is then used to detect quadrotor propellers. Other works such as \cite{burusa2022attention} and \cite{Thanasis,freeman20233d,khaki2021deepcorn} use synthetic data to perform tomato plant and corn field reconstruction, respectively.

To the best of our knowledge, the first major breakthrough in applying synthetic data for deep learning-based fruit counting was in \cite{rahnemoonfar2017deep}, where red circles of various sizes were scattered onto a blurred green and brown background in order to count tomatoes. This novel approach resulted in approximately 91\% counting accuracy and inspired many other synthetic data approaches. Synthetic data is not just successful in fruit counting, but also in detecting oysters. Lin \textit{et al.} \cite{lin2022oystersim} propose a novel open-source simulation that is used to generate photo-realistic synthetic images of oyster reefs. Using a combination of the synthetically generated oyster reef images and read images, Lin \textit{et al.} \cite{OysterNet} trains a semantic segmentation model resulting in a major improvement than when only trained on real data. Another synthetic data generation approach for precision agriculture is presented by Blekos \textit{et al.} \cite{blekos2021analysis} where a synthetic olive tree (not containing olives) is designed and used in detecting \textit{Verticillium Wilt}, a fungal disease that can lead to the death of an olive tree. In our paper, we create a lightweight synthetic olive tree, design a novel olive model, and train a deep semantic segmentation model for olive detection, seeing an improvement when real data is augmented by synthetically generated data. 

%first paragraph: overview of all related works that I will cover in categories. 2nd paragraph - end: For each category, identify the problem and possible solutions by authors that are grouped by similarity. "This problem was addressed via this method by [5-8]." For the problems that are similar to mine, identify the differences between theirs and mine and why my problem is more difficult.

%This is a complex task because unlike other fruits (apples, oranges, tomatoes, berries, etc.), the olive appears in a range of numerous colors at the time of harvest, making detection much more difficult

%divide this into 3 sections: precision agriculture for counting (minor), precision agriculture for yield estimation (major) and synthetic data (minor)

%[Cite Tejada/Zervakis] uses hyperspectral and multispectral sensing onboard a UAV approach to distinguish grove diseases. [Cite Thanasi and all other reconstruction efforts] In the U.S., farmers use it to autonomously learn factors about their crops such as soil acidity, plant hydration, and harvest estimations.

%% file: sections/3.proposed_method.tex
\section{Synthetic Olive Tree}
\label{section:Synthetic_Olive_Tree}
Given the arduous nature of real-world data collection and labeling for semantic segmentation, we propose a system to rapidly streamline the process of creating a rich, diverse hybrid dataset containing real and synthetic data. The need for such a swift system is highlighted further, given that each olive tree contains around 2500 to 17500 olives, on average \cite{dz_2022}. The resulting generated synthetic data is then combined with real data to train a segmentation model to detect olives as shown in Fig.~\ref{fig:overview}. In Sec.~\ref{3D Model}, we illustrate the process of creating synthetic data by modeling the geometry of an olive tree utilizing Blender\texttrademark, a 3D open-source graphics software to perform rendering \cite{blender}. We also ensure that our 3D model is light enough to be rendered under constrained computational resources (such as a modern-day laptop), further democratizing our synthetic data pipeline. In Sec.~\ref{translation}, we explain how modern image-to-image translation methods may be applied to make the rendered synthetic images appear photorealistic.
\begin{figure}[t]
\includegraphics[width=0.8\linewidth]{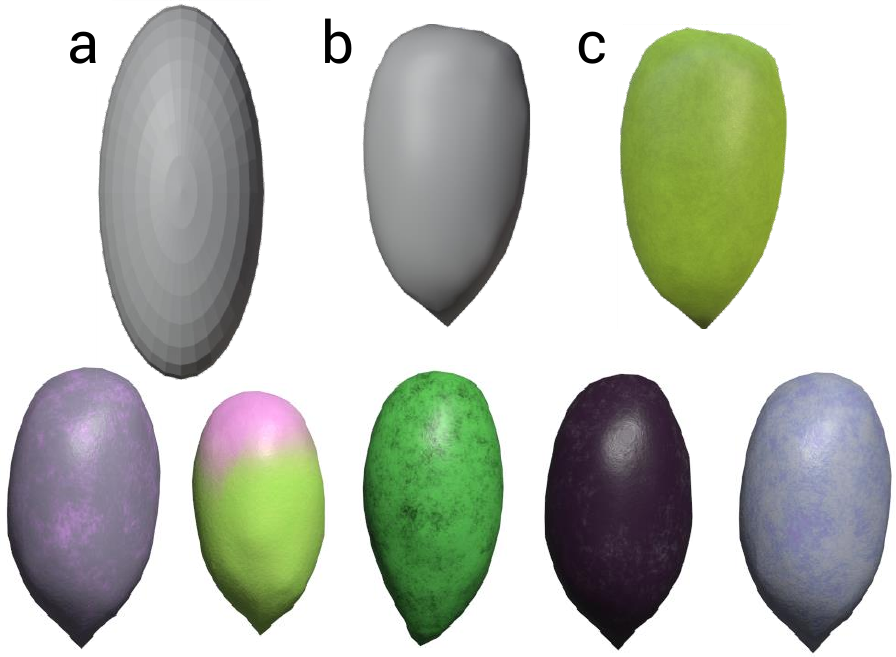}
\centering
\caption{(a) the 3D modeling of our ellispoid. (b) the 3D modeling of an olive (c) an example olive rendered with texture.}
\label{fig:Olive_models}
\end{figure}

\subsection{Olive Tree 3D Model}
\subsubsection{Synthetic Olive}
\label{3D Model}
To generate synthetic data for an olive tree, we first need to create its most fundamental component: the olive.  We model the geometry of an olive, by using an ellipsoid mesh as a baseline:

%needs to be 3 separate lines
% \begin{align*} 
%     x &=  u\sin(v) \\ 
%     y &= u\cos(v) - sin(v)\\
%     z &= u\cos(v) \numberthis \label{eqn}
% \end{align*}
\begin{equation}
    \frac{x^2}{a^2}+ \frac{y^2}{b^2} + \frac{z^2}{c^2} = 1
\label{eq:elipsoid}
\end{equation}
Here in Eq.\ref{eq:elipsoid}, $a=1.2\ \text{cm}$ and $b=1.2\ \text{cm}$ are the width factor of the ellipsoid while $c=2.3\ \text{cm}$ is the length and width of the olive as shown in Fig.~\ref{fig:Olive_models}\textcolor{red}{a}. We added some randomness in the scaling factor to make it more realistic. Then, the vertices near the end of the ellipsoid are extended to create a sharp point, matching the shape of a real olive.

% \begin{python}
% import numpy as np
% import matplotlib.pyplot as plt

% def plot_olive():
%     # Define the parameters of the olive shape
%     a, b, c = 0.9, 1.3, 1.6
%     t = np.linspace(0, 2*np.pi, 20)
%     x = a*np.sin(t)
%     y = b*np.cos(t) - np.sin(t)
%     #z = c*np.cos(t)

%     # Create a 2D plot
%     fig, ax = plt.subplots()

%     # Plot the olive in the X-Y plane
%     ax.plot(x, y, color='green')

%     # Add some finishing touches
%     ax.set_xlabel('X')
%     ax.set_ylabel('Y')
%     ax.set_title('Olive')

%     plt.show()
% \end{python}

% \begin{tikzpicture}
%     \begin{axis}[
%         xlabel={$x$},
%         ylabel={$y$},
%         title={Olive},
%         width=10cm,
%         height=10cm,
%         axis equal,
%         axis lines=middle,
%         xmax=2,
%         xmin=-2,
%         ymax=2,
%         ymin=-2,
%         tick style={draw=none}
%         ]
%         \addplot[color=green, domain=0:2*pi, samples=20] ({0.9*sin(deg(x))},{1.3*cos(deg(x)) - sin(deg(x))});
%     \end{axis}
% \end{tikzpicture}

% Here, $h=2.4\text{cm}$ and $k=0.6\text{cm}$ are the height and heigh-scaling factor of the ellipsoid while $b=1.3\text{cm}$ is the length and width of the olive. ($z \leq k\sqrt{x^2 + y^2}$) accounts for the pyramid bottom of the olive, while ($z > k\sqrt{x^2 + y^2}$) represents the ellipsoid part of the olive. The two parts of the function are smoothly interpolated at the boundary $z=k\sqrt{x^2 + y^2}$ to create a smooth transition between the pyramid and the ellipsoid. %Then, the vertices near the bottom edge are extended to match the shape of a real olive. 
Subsequently, a smoothing process is applied to the olive 3D model to eliminate any form of rigidness. The final dimensions of our 3D olive model are $2.5\text{cm} \times 1.4\text{cm} \times 1.4\text{cm}$ respectively as such are the approximate dimensions of olives when ready for harvest. Fig.~\ref{fig:Olive_models}\textcolor{red}{b} demonstrates the process of modeling our proposed olive models.
% \begin{figure}[b]
% \includegraphics[width=0.8\linewidth]{./figures/draft_olives.pdf}
% %\includegraphics[width=\textwidth,height=\textheight,keepaspectratio]{sequence.png}
% \centering
% \caption{A zoomed-in look at the different olives we model.\textcolor{red}{I don't know shall we combine Fig 4 and 5. kinda save the same purpose of showing 3D models of the olive. }}
%\label{fig:Olive_models_with_different_colors}
%\end{figure}
We then initiate the process of adding texture to the olive model. The appearance of an olive is impacted by the lighting, shading, and albedo of the environment. All olives begin as green fruits and gradually transform into darker shades, ending in an almost black, dark purple color as the harvest season progresses (see Fig. \ref{fig:Olive_models}\textcolor{red}{c} and its examples underneath). 
%We model such differences in textures as shown in figure [make a figure of real olives and various colors] where in the process of changing color, olives may appear as a lighter green, light purple, orange, burgundy, blue, etc. 
Occasionally, a single olive may have two different colors simultaneously. In addition, olives usually do not have a smooth texture but may exhibit bumps and wrinkles on their surface. Given the grove environment, it is possible for olives to accumulate dust, providing a lighter shade on certain sections of the surface. Finally, many olive groves suffer from the insect \textit{dacus olea}, which punctures the olive causing black spot(s) to appear on its surface. We consider all the above variations when designing the texture of an olive. We model a set of eleven unique textures to be applied onto the 3D olive model, each varying in noise, roughness, distortion, bump, and color mixing. This ensures that the set of synthetic olives closely resembles real-world olives.  
\subsubsection{Olive tree and leaves}
\begin{figure}[t]
\includegraphics[width=1.0\linewidth]{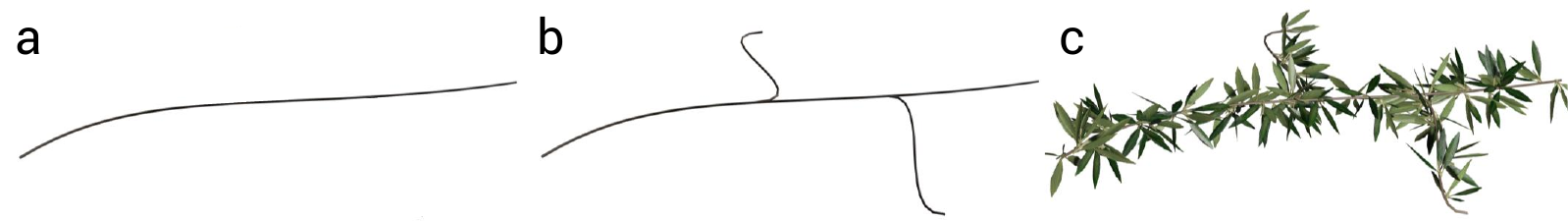}
\centering
\caption{Fig. 5. (a) a single bezier curve with 2 control points. (b) the bezier curve we use as a branch. (c) the bezier curve with 3D-scanned leaves scattered onto it.}
\label{fig:branch_models}
\end{figure}

Next, we create the leaves and the branches of the olive tree. We scan a pair of leaves from a real olive tree to model the 3D geometry. The scanner automatically provides the texture of an olive leaf. To model the branches and subbranches, we utilize the bezier curve as a baseline, and is modeled as follows:

\begin{equation}
B(t) = \sum_{i=0}^n {n \choose i}(1-t)^{n-i}t^iP_i
\label{eq:bezier}
\end{equation}

where $t$ represents the position of the point on the curve and $P_i$ represents the $i^{th}$ control point. In Eq.~\ref{eq:bezier} we select $n=5$ control points per branch. We then prune the ends of the curves in Blender to make them more realistic. To match the texture of an olive tree's branches onto our 3D model's branches, we use a dark brown as our base color and apply dark noise to the branches. This process is illustrated in Fig. \ref{fig:branch_models}.

In an effort to make our 3D model more accessible to those without high-end computing devices, we decide to make a lighter version of an olive tree model without a trunk or a dense canopy. \emph{Choosing to do so drastically reduces the average time needed for rendering synthetic image-mask pairs.} First, we eliminate the trunk. Then, we reduce the number of leaves and branches required by using an invisible 2D plane as a baseline for their scattering. Finally, to address the excessively high number of vertices on the 3D-scanned leaf, we apply a limited dissolve onto the leaf model for simplification, further democratizing (making it easily accessible for all) our 3D model.

To complete the model, we first create four separate layers of  scattered leaves, two separate layers of  scattered branches, and one layer of scattered olives per rendering session in the formation of 2D planes. The randomness of leaf orientation is set to no more than 9\degree and the size randomness is set to be no more than 10\% of the original dimensions. When scattering the olives onto the leaves and branches, we place them behind one leaf layer in order to simulate the obscurity that real olive leaves provide to the olive fruit. We apply orientation randomness within 45\degree and size randomness within 5\% of the original dimensions to all olives. In total, we scatter approximately 2,600 olives onto the leaves per rendering session. In total, we run 16 rendering sessions to produce 15,960 image-mask pairs. Example renderings are shown in Fig.~\ref{fig:translation}\textcolor{red}{c}.
Since the Mediterranean olive harvest occurs during the months of mid-Autumn, we emulate appropriate background and lighting conditions for our 3D model. We first simulate dry, sunny days with little to no clouds using Blender's sky, Musgrave, and gradient texture nodes. We then simulate a cloudy, overcast environment using Blender's dynamic sky. Finally, we place a plane behind the final leaf layer with ground texture projected onto it via UV mapping to simulate a UAV flying above a canopy with the ground as the background. We also UV-map an image of scattered leaves onto a background plane to simulate a UAV flying at the height of the canopy.

\subsubsection{Rendering}
At this point, the 3D olive tree model is ready to be rendered into image-mask pairs, a very computationally demanding task. Therefore, we take the following steps to shorten the rendering process, expanding the dataset's availability to computers with lower processing power (such as modern laptops). \begin{itemize}
    % \item We select Cycles, a more computationally expensive but accurate ray-tracing method
    \item We utilize Intel OpenImageDenoise \cite{intel}, an open-source library of denoising filters made for images rendered with ray tracing
    \item We lower the maximum number of samples in Cycles render (number of paths to trace for each pixel in the final render) to 50, still producing a high-quality output
    \item We eliminate all subsurface scattering in all objects
\end{itemize}

\subsection{Image-to-Image Translation}
\label{translation}
Image-to-image translation is a technique that converts an image from a source domain to a target domain (e.g. convert an image of a horse to a zebra) \cite{CycleGAN2017} \cite{pix2pix2017}. Typically, synthetic data does not possess the realistic features of the real data domain. Models trained on untranslated synthetic data perform poorly on real-world testing data. Therefore, we conduct image-to-image translation to transform synthetically rendered images in Blender of our 3D model to the real-world olive tree domain. The objective is that training with synthetic data closely resembling the real world will result in higher accuracy models. Specifically, we employ Unpaired Image Translation via Vector Symbolic Architectures (VSAIT) \cite{theiss2022unpaired} to translate the Blender-generated images to the real-world domain. We also choose VSAIT in order to reduce occurrences of semantic flipping (when a green olive gets translated into a leaf, corrupting the translated data) given that VSA-based methods are capable of learning high-level, abstract concepts \cite{SparseDistributedMemory}.
%We choose VSAIT as it makes use of Vector Symbolic Architectures (VSA), a new paradigm in machine learning using high-dimensional vectors to represent all entities. 
 
Formally, VSAIT is trained to learn the translation between the Blender-generated synthetic olive tree images $X$ to the real olive grove images $Y$. The three major components of this network are the Generator $G$, the Discriminator $D_Y$, and the Source $\leftrightarrow$ Target Mapper F. The overall loss consists of the sum of the Hypervector Adversarial Loss and the VSA-based Cyclic Loss \cite{theiss2022unpaired}.

\textbf{Hypervector Adversarial Loss:} This ensures the similarity between the synthetic translated image hypervectors $v_{G(X)}$ and the real target hypervectors $v_Y$. This is done by applying the VSA binding operation on hypervector mapping $F(v_x)$ and source vectors $v_X$ to yield a hypervector of synthetic features mapped to the real domain. The Hypervector Adversarial Loss is calculated as:
% \begin{equation}
%         L_{GAN}(G,D_Y,X,Y) = \mathbb{E}_{y \sim pY(y)}[logD_Y (v_y)]\\
%                             +\mathbb{E}_{y \sim pY(y)}[logD_Y (v_y)]\\
%                             +\mathbb{E}_{y \sim pY(y)}[logD_Y (v_y)]
% \end{equation}
%Ask Xiaomin how to get this as equation 2
\begin{align*}
%\label{eq:pareto_mle2}
 \mathcal{L}_{GAN}&(G,D_Y,X,Y)=    \mathbb{E}_{y \sim pY(y)}[\log D_Y (v_y)]\\
                & + \mathbb{E}_{x \sim pX(x)}[\log(1-D_Y (v_{G(x)}))]\\
                & + \mathbb{E}_{x \sim pX(x)}[\log(1-D_Y (v_x \otimes F(v_x)))] 
                \numberthis \label{eq:pareto_mle2}
\end{align*}

\textbf{VSA-based Cyclic Loss:} This cyclic loss minimizes occurrences of semantic flipping by constraining $G$ so that similar hypervectors may be returned when mapping translated vectors back to the synthetic domain from the real domain. Meaning, $v_X \approx v_{G}(X)\rightarrow X$. The VSA-based Cyclic Loss is calculated as: 
\begin{equation}
        \mathcal{L}_{VSA}(G,X) = \mathbb{E}_{x \sim pX(x)}
        \left[ \frac{1}{n}\sum_{i=1}^{n}dist(v^i_x,v^i_{G(x) \rightarrow x})\right]
\end{equation}

where the \textit{dist} is the cosine distance between the source hypervectors and inverted translated hypervectors.

%need help to get this right.
Therefore, the overall loss is as follows:
\begin{equation}
         \mathcal{L}(G,D_Y,X,Y) = \mathcal{L}_{GAN}(G,D_Y,X,Y)+\mathcal{L}_{VSA}(G,X)
\end{equation}

\begin{figure}[t]
\includegraphics[width=\linewidth]{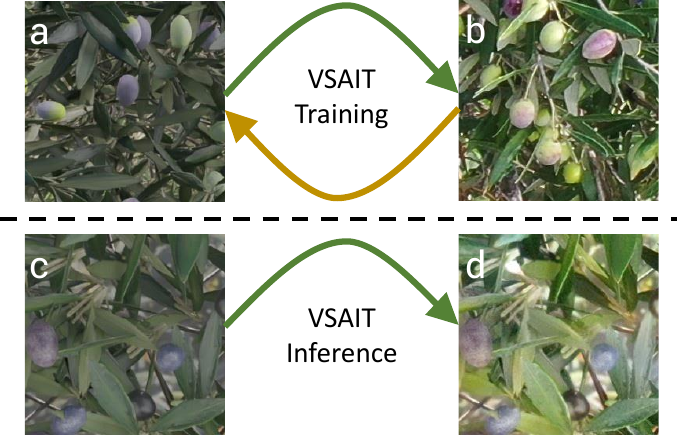}
\centering
\caption{Image-to-image translation (a) A single sample of the simulated olive, (b) A single sample of the real olive, (c)
Simulated olive image, (d) Synthetic olive image translated into real
world for photorealism.}
\label{fig:translation}
\end{figure}

To create $X$, the source domain, we render 12,768 images of our 3D olive tree model for training and 3,192 images as validation. To create $Y$, the target domain, we use 12,452 UAV-captured images for training and 3,113 for validation. Overall, we train VSAIT for 98 epochs using its default settings (see Fig.~\ref{fig:translation}\textcolor{red}{a},~\ref{fig:translation}\textcolor{red}{b}). Then, we perform inference to apply VSAIT to rendered images (Fig.~\ref{fig:translation}\textcolor{red}{c}), yielding a photorealistic synthetic output (Fig.~\ref{fig:translation}\textcolor{red}{d}).

We ensure that the images used in the $Y$ domain for image-to-image translation do not overlap with any of the real images that will later be used in testing segmentation models. When $X$ gets translated to $Y$, it learns the features of $Y$. Training a segmentation model using synthetic images translated to the same $Y$ as the real-world test set is just like training with testing data as input (and must be avoided). 

\subsection{Synthetic Data Color Input Space}
After the synthetic data undergoes image-to-image translation, the olives in the synthetic data still stand out brightly, unlike in real olive groves. Furthermore, the  synthetic data translated by VSAIT can easily still be distinguished from images of the real domain, if looked at closely. Therefore, in an attempt to make the olives in the VSAIT output less distinguishable (just like in the real-world), we create a new input space, with the goal of making the olives more concealable.  We draw inspiration from \cite{yu2022udepth} who propose a novel input space from  ($\text{RGB} \rightarrow \text{R, M=max(G,B), I}$) where I is the overall intensity. However, Yu \textit{et al.}\cite{yu2022udepth} apply this technique to the underwater domain where there is a deficiency of red reflections %I think that's correct but I haven't checked
to extract a clearer image for depth estimation. We propose a color input space for the opposite purpose: to obscure the olives further into their backgrounds. \emph{This new input space may be applied to all synthetic agricultural domains with the purpose of blending any small fruit into its respective backgrounds} and is as follows (called IGA):
$$ RGB \rightarrow (I,G,Avg(R,B))$$
Converting the VSAIT-translated synthetic data to this input space yields a slight improvement as shown in Table \ref{tab:table_label} and is labeled as $O_{IGA\_and\_real}$.

%% file: sections/4.Experiments_And_Results.tex
\section{Experiments and Results}
\label{section:Experiments_and_results}
We first describe our datasets. Then we analyze our segmentation experiments on models trained on images with and without synthetically generated data.
\subsection{Dataset Overview}
%We describe the synthetic dataset and then we describe the process of creating the real dataset.
\noindent
\textbf{Synthetic Data:}
As described in Section~\ref{section:Synthetic_Olive_Tree}, we initiate multiple rendering processes of our Blender scene, each reproducing parts of an olive tree with a blue sky, leaf-only, or ground texture background. In addition, we utilize multiple lighting conditions to augment the variety of our synthetic data. Afterward, we initiate the image-to-image translation process \cite{theiss2022unpaired}. The result of these two processes is the speedy generation of 15,960 synthetic image-mask pairs.

\noindent
\textbf{Real Data: }
We conducted several UAV flights over the island of Crete, Greece. Our UAV is the DJI Mavic 2 Pro with a Hasselblad L1D-20C camera. To make the dataset as diverse as possible, the flights were staggered to record differences in cloud cover and albedo, and to capture images of olives during different temporal stages of the olive harvest, approximately ranging from one to two months. We also chose to only set our ISO to 100 for daytime capture and 400 for evening capture. We then labeled our UAV-captured images by hand using LabelMe \cite{Wada_Labelme_Image_Polygonal}. In total, we have 3,113 real-world image-mask pairs in patches of size $ 256 \times 256 \times 3$ and $ 256 \times 256 \times 1$ and we labeled approximately 2,600 olives. The maximum number of olives per unpatched image is around 600.

\subsection{Experiments}
We construct two training sets: one with 15,960 synthetic and 100 real image-mask pairs and another with only 100 real image-mask pairs. This was decided to reflect the reality that it is very difficult to manually create a large enough labeled olive dataset. We select UNet \cite{unet} as our model architecture and train with several backbones \cite{resnet,efficientnet} as shown in Table \ref{tab:table_label}. To standardize experiments across our two data sets, we only employ the Adam optimizer \cite{adam} with an initial learning rate of 0.001 and use the Jaccard index \cite{jaccard} as our loss function. All models are trained for a maximum of 100 epochs using a batch size of 32. We use minimal real-world data as it is arduous to obtain and label, simulating situations where one may only have limited access to real data, which are more realistic cases. 

Our testing set consists of more than 3000 real-world image-mask pairs that are hand-labeled \cite{Wada_Labelme_Image_Polygonal}. Our metric of choice is the Intersection over Union, as there is only one class of objects to segment. 
We opt out of using mean Average-Precision as an evaluation metric given that models trained on mostly synthetic data perform well only with low-confidence score thresholding, as is in our case. 

\subsection{Results}
Based on our experiments, as shown in Table. \ref{tab:table_label}, we observe that the models with identical architectures and backbones trained on both synthetic and real images outperform the baseline models, which are trained without synthetic data. In our \emph{Training Data} column, $O_{real}$ represents the set of just real-world training data, $O_{syn\_and\_real}$ represents the set of real-world and VSAIT-translated synthetic data, and $O_{IGA\_and\_real}$ represents the VSAIT-translated synthetic data converted to the IGA input space. Our biggest marked improvement, when compared to the baseline, was observed from 24.22\% to 40.22\% IoU, corresponding to a percent change of 66\% when using ResNet101 \cite{resnet} as our backbone. We also observe that when using EfficientNetB5 \cite{efficientnet}, ResNet101, and ResNet152, the highest-performing models are trained on VSAIT-translated images converted to the IGA input space. Fig.~\ref{fig:detection_result} shows the difference in segmentation prediction when using models trained with and without synthetic data. When using an EfficientNetB5 backbone, we observe a 12.64\% jump in IoU over the baseline when the model is supplemented with synthetic data in the IGA input space. Similarly, when we use Resnet152 as a backbone, we observe an increase of 16.34\% compared to the baseline when supplemented with unedited VSAIT-generated data.
\subsection{Discussion} Detecting olives in an olive grove is a problem in the low training data regime domain. Under this setting, it is not convenient to label vast amounts of olive tree images and hence synthetic data offers a reasonable alternative. In this paper, we simulate such a situation and have demonstrated that supplementing a low amount of real data with easy-to-generate synthetic images improves the segmentation results by 66\% as shown in Table \ref{tab:table_label}.

% \begin{equation*}
% \mathrm{IoU} = \frac{\vert y_\mathrm{truth} \cap y_\mathrm{pred} \vert}{\vert y_\mathrm{truth} \cup y_\mathrm{pred} \vert} = \frac{\sum \limits_{i=1}^n \mathbb{I}(y_{\mathrm{truth},i} = olive \land y_{\mathrm{pred},i} = olive)}{\sum \limits_{i=1}^n \mathbb{I}(y_{\mathrm{truth},i} = olive \lor y_{\mathrm{pred},i} = olive)}
% \end{equation*}

%$$\mathrm{IoU} = \frac{\vert A \cap B \vert}{\vert A \cup B \vert}$$

% \begin{table}[h]
% \centering
% \begin{tabular}{|c|c|c|c|}
% \hline
% \textbf{Data Type} & \textbf{Model} & \textbf{Backbone} & \textbf{IoU} \\
% \hline
% Data 1 & Data 2 & Data 3 & Data 4 \\
% Data 5 & Data 6 & Data 7 & Data 8 \\
% Data 9 & Data 10 & Data 11 & Data 12 \\
% \hline
% \end{tabular}
% \caption{I have many more experiments to run. This is nowhere near enough.}
% \label{tab:table_label}
% \end{table}

\begin{table}[t!]
\caption{IoU using real only, synthetic/real, and synthetic IGA/real} 
\centering
\begin{tabular}{||l|l|c||}
\hline
     \textbf{Training Data} & \textbf{Backbone} & \textbf{IoU} \\
\hline 
    $O_{real}$ &  EfficientNetB5 & 41.41\% \\
    $O_{syn\_and\_real}$  & EfficientNetB5 & 53.06\% \\
    $O_{IGA\_and\_real}$ & EfficientNetB5 & \textbf{54.05\%} \\
\hline 
     $O_{real}$ &  ResNet101 & 24.22\% \\
     $O_{syn\_and\_real}$  & ResNet101 & 36.28\% \\
    $O_{IGA\_and\_real}$ & ResNet101 & \textbf{40.22}\% \\
\hline
    $O_{real}$ &  ResNet152 & 28.99\% \\
     $O_{syn\_and\_real}$  & ResNet152 & 40.54\% \\
    $O_{IGA\_and\_real}$ & ResNet152 & \textbf{45.33}\% \\
\hline
\end{tabular}
\label{tab:table_label}
\end{table}
%mAP measures diffent confidences, however, the model trained on synthetic only performs well on low confidence scores, making the metric of less significance. doesnt make sense to measure high cinfidence.

%\begin{tabular}{||c|c||}
%\hline
%     \textbf{Model Architecture} & \textbf{IoU} \\
%\hline \\
%    only real data & 41\% \\
%\hline \\
%    syn + real data IGavg & 54\% \\
%\hline
%\end{tabular}

%% file: sections/5.conclusions.tex
\section{Conclusion and Future Work}
\label{section:Conclusion}
In this paper, we presented a novel approach to detect olives in an olive grove without the need to manually label data. 
%We chose to generate synthetic data, eliminating the one must collect and label several thousands of very dense olive grove images, a manual, arduous, and daunting task. This paper presents a novel approach to detect olives . It also presents the world’s first olive detection dataset consisting of mostly synthetic and some real images of olive trees. 
We described a technique to generate a photorealistic 3D model of an olive tree. Its geometry is then simplified for lightweight rendering purposes, thus democratizing the synthetic dataset. The goal is to supplement low levels of real data with our synthetic data to improve detection results. To evaluate our approach, we conducted experiments exhibiting promising results with a maximum observed change in IoU of 66\%, when using our proposed method versus low-data regime methods. To conclude, this demonstrates that when access to real, labeled data is restricted or non-attainable, a combination of mostly synthetic data and real data can help to enhance olive detection.